\documentclass[lettersize,journal]{IEEEtran} 
\usepackage{amsmath,amsfonts} 
\usepackage{array} 
\usepackage[caption=false,font=normalsize,labelfont=sf,textfont=sf]{subfig} 
\usepackage{textcomp} 
\usepackage{stfloats} 
\usepackage{url} 
\usepackage{verbatim} 
\usepackage{graphicx} 
\usepackage{cite} 
\usepackage{caption}
\usepackage{amsthm}
\usepackage{cases}
\usepackage[english]{babel}
\usepackage[ruled]{algorithm}   % 使用规则样式
\usepackage{algpseudocode}      % 伪代码，避免与 algorithmic 冲突
\usepackage{multirow}
\usepackage{array}
\usepackage{bm}
\usepackage{cite}
\usepackage{stfloats}
\usepackage{verbatim}
\usepackage{amsfonts,amssymb}
\usepackage{array}
\usepackage{color}
\usepackage{booktabs}
\usepackage{float} 

\graphicspath{{images/} }

\allowdisplaybreaks

\renewcommand{\baselinestretch}{1.0}  % 确保单倍行距

\makeatletter

\newcommand{\Rmnum}[1]{\expandafter\@slowromancap\romannumeral #1@}
\makeatother

\ifCLASSINFOpdf
\else
\fi

\DeclareCaptionLabelFormat{fig}{Fig.~#2}
\begin{document}
	
	\title{
		Semantic Communication-Enhanced Split Federated Learning for Vehicular Networks: Architecture, Challenges, and Case Study
	}
	\author{Lu Yu, Zheng Chang,~\IEEEmembership{Senior Member, IEEE,} and Ying-Chang Liang~\IEEEmembership{Fellow,~IEEE}
		
		\thanks{L. Yu and Z. Chang are with School of Computer Science and Engineering, University of Electronic Science and Technology of China, Chengdu 611731, China. Z. Chang is also with Faculty of Information Technology, University of Jyv\"askyl\"a, FIN-40014 Jyv\"askyl\"a, Finland. Y-C. Liang is with Center for Intelligent Networking and Communications (CINC), University of Electronic Science and Technology of China, 611731 Chengdu, China. }
		
		\thanks{Accepted for publication in IEEE Communications Magazine.}
	}
	%email zheng.chang@jyu.fi. richard.ndjiongue@wits.ac.za, ycliang@ieee.org
	
	% The paper headers
	\markboth{Journal of \LaTeX\ Class Files,~Vol.~14, No.~8, August~2024}%
	{Shell \MakeLowercase{\textit{et al.}}: A Sample Article Using IEEEtran.cls for IEEE Journals}
	\maketitle

%	\IEEEpubid{
%		\begin{minipage}{\textwidth}
%			Accepted for publication in IEEE Communications Magazine.
%	\end{minipage}}
%	\IEEEpubidadjcol

	\begin{abstract}
		Vehicular edge intelligence (VEI) is vital for future intelligent transportation systems. However, traditional centralized learning in dynamic vehicular networks faces significant communication overhead and privacy risks. Split federated learning (SFL) offers a distributed solution but is often hindered by substantial communication bottlenecks from transmitting high-dimensional intermediate features and can present label privacy concerns. Semantic communication offers a transformative approach to alleviate these communication challenges in SFL by focusing on transmitting only task-relevant information. This paper leverages the advantages of semantic communication in the design of SFL, and presents a case study the semantic communication-enhanced U-Shaped split federated learning (SC-USFL) framework that inherently enhances label privacy by localizing sensitive computations with reduced overhead. It features a dedicated semantic communication module (SCM), with pre-trained and parameter-frozen encoding/decoding units, to efficiently compress and transmit only the task-relevant semantic information over the critical uplink path from vehicular users to the edge server (ES). Furthermore, a network status monitor (NSM) module enables adaptive adjustment of the semantic compression rate in real-time response to fluctuating wireless channel conditions. The SC-USFL framework demonstrates a promising approach for efficiently balancing communication load, preserving privacy, and maintaining learning performance in resource-constrained vehicular environments. Finally, this paper highlights key open research directions to further advance the synergy between semantic communication and SFL in the vehicular network.
	\end{abstract}
	
	\IEEEpeerreviewmaketitle

	\section{Introduction}
	The advent of autonomous driving and vehicular networking technologies heralds a transformative shift in transportation, promising substantial improvements in road safety, traffic efficiency, and overall mobility. This progression towards highly intelligent transportation systems (ITS) necessitates real-time, robust data processing and sophisticated decision-making. Vehicular edge computing (VEC) and the subsequent vehicular edge intelligence (VEI)—integrating AI at the network edge—are foundational for these advancements, empowering vehicles with complex ITS functionalities. However, conventional centralized learning (CL) approaches for VEI face significant impediments: substantial communication overhead, prohibitive latency, considerable privacy risks with sensitive data transmission \cite{VEN_data}, and scalability limitations in dynamic vehicular networks. \par
	
	To overcome CL's deficiencies, distributed learning (DL) is essential for enabling VEI, mitigating high communication costs and privacy vulnerabilities via collaborative on-device training. Key DL evolutions include federated learning (FL), which introduces local training with server-side parameter aggregation, reducing raw data transmission but imposing computational demands on vehicular users (VUs) and facing communication inefficiencies with large models \cite{FL_Pruning}. Split learning (SL) lessened VU computation by partitioning models between VUs and VEC server, but leads to frequent intermediate data (``smashed data") transmission, potential privacy leakage, and sequential training inefficiencies. Split federated learning (SFL) then emerges, hybridizing FL's parallel training with SL's reduced client computation \cite{VEN_SFL_2}. While powerful, SFL also faces several challenges: significant communication overhead from transmitting high-dimensional intermediate features and, depending on its configuration, potential label privacy concerns. Addressing these SFL challenges is crucial for effective VEI deployment. \par
	
	The efficacy of SFL frameworks is further constrained by traditional communication protocols, ill-suited for dynamic, resource-limited vehicular networks. This leads to inefficiencies like excessive overhead from transmitting voluminous, sometimes irrelevant, intermediate SFL features, sensitivity to fluctuating network conditions, and suboptimal resource use due to treating all data bits equally. In response, semantic communication has garnered attention, shifting focus from bit-perfect reconstruction to conveying task-relevant meaning \cite{SC_1}. For SFL in vehicular networks, semantic communication promises drastically reduced overhead by transmitting only critical information, thereby improving efficiency and robustness against adverse channel conditions.\par
	
	The convergence of semantic communication with SFL offers a potent solution to these communication limitations, enabling more efficient data exchange for faster training and improved responsiveness in VEI systems by transmitting only semantically relevant aspects of intermediate features. While generally enhancing SFL, applying semantic communication to specialized SFL architectures designed for challenges like enhanced privacy is insightful. U-Shaped SFL (U-SFL), an SFL configuration retaining initial (head) and final (tail) model layers on the client, is one such specialization that inherently enhances label privacy by localizing sensitive information and loss computation \cite{USFL-Original}. Building on this, this paper also introduces a semantic communication-enhanced U-Shaped split federated learning (SC-USFL) framework. SC-USFL serves as a comprehensive case study demonstrating how task-oriented semantic communication can be deeply integrated into an advanced, privacy-aware SFL architecture like U-SFL, thereby harnessing synergistic benefits for vehicular networks. \par
	
%	In this work, we are motivated by the pressing need to overcome the communication bottleneck in SFL for VEI systems, especially in scenarios where data privacy is paramount. Recognizing the immense potential of semantic communication, this paper presents a comprehensive tutorial to demonstrate how this emerging paradigm can be deeply integrated into an advanced, privacy-preserving SFL architecture. To the best of our knowledge, this work represents an early attempt to provide a holistic framework and performance analysis for an adaptive, semantic-enhanced SFL system tailored for dynamic vehicular networks \cite{FL_semantic_2}.\par
	
%	{\color{red}
%		In this work, we propose the SC-USFL framework to address the critical bottlenecks of communication efficiency and data privacy in VEI. Distinct from generic SFL approaches that transmit raw intermediate features \cite{FL_semantic_2} and often compromise label privacy, this work introduces a novel architectural synergy by deeply integrating a DeepJSCC-based semantic communication module (SCM) into a U-Shaped split topology. This design not only minimizes bandwidth consumption by transmitting task-relevant semantics but also structurally guarantees label privacy by anchoring the classification tail locally. Furthermore, to handle dynamic vehicular environments, we design a network status monitoring (NSM) that introduces channel-aware adaptivity, allowing the system to dynamically adjust the semantic compression ratio based on real-time channel conditions. \par
%	}
	
	In this work, we propose the SC-USFL framework to address VEI communication and privacy bottlenecks. Distinct from generic SFL \cite{FL_semantic_2}, SC-USFL integrates a deep joint source channel coding (Deep JSCC)-based semantic communication module (SCM) into the U-Shaped topology. This synergy minimizes bandwidth via task-oriented compression while structurally guaranteeing label privacy by anchoring the classification tail locally. Additionally, we introduce a network status monitor (NSM) to enable channel-aware adaptivity, dynamically adjusting compression ratios to maintain robustness in dynamic vehicular environments.  \par

	The remainder of this article is organized as follows. We first delve into the convergence of distributed learning and semantic communications for vehicular edge intelligence, outlining their evolution and inherent challenges. Then, we introduce our proposed SC-USFL framework as a detailed case study, including its core SCM and NSM components. Subsequently, we provide a comprehensive performance evaluation of the SC-USFL framework. Finally, we explore open research directions in this burgeoning field and conclude the paper with key insights and future prospects.\par

	\section{Distributed Learning and Semantic Communications for VEI: Convergence and Challenges}
	The imperative to overcome centralized learning's limitations in VEI has spurred the evolution of DL architectures. This progression has sought to balance data privacy, computational load distribution, and communication efficiency.

	\subsection{Distributed Learning: Architectural Evolution and Communication Bottlenecks in VEI}
\begin{figure}[t]
		\centering
		\renewcommand{\figurename}{Fig.}
		\includegraphics[width=0.7\columnwidth]{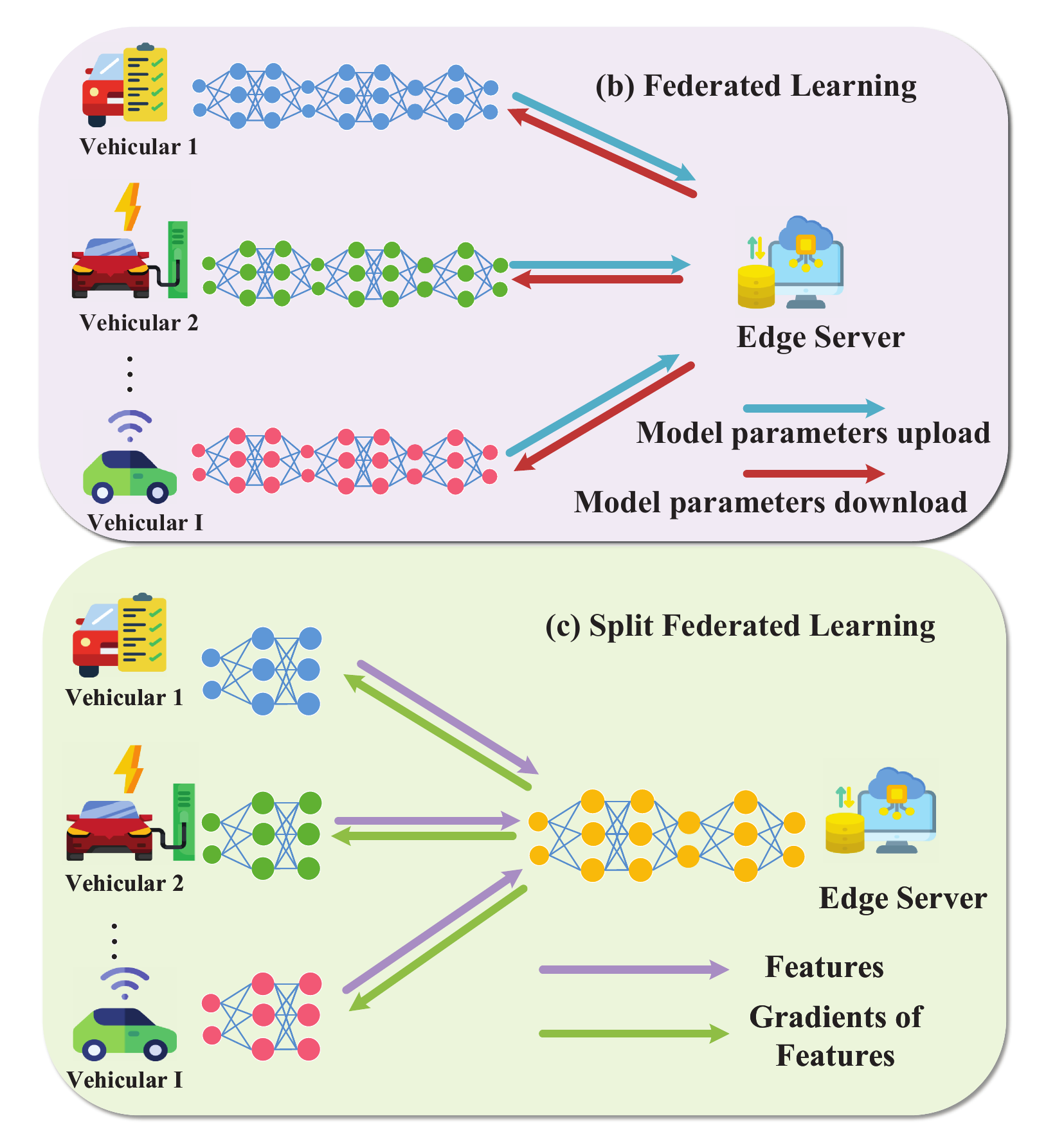}
		\caption{The workflow of (a) federated learning and (b) split federated learning.}
		\label{fig:DistributeLearning_workflow}
	\end{figure}

	To balance privacy and efficiency, distributed architectures have evolved from centralized paradigms. \textbf{FL} (as illustrated in Fig.~\ref{fig:DistributeLearning_workflow}(a)) enables VUs to train models locally and aggregate parameters at a server, ensuring raw data privacy. However, FL's local training is computationally intensive for VUs, and transmitting large model updates incurs significant overhead, especially under dynamic vehicular topologies. To alleviate client-side load, \textbf{SL} partitions models into client and server segments. While SL reduces VU computation, it introduces frequent transmission of high-dimensional ``smashed data", sequential training bottlenecks in large networks, and potential privacy leakage from intermediate features or labels.
	
	\textbf{SFL} was proposed to hybridize FL's parallel training with SL's reduced client load, as illustrated in Fig.~\ref{fig:DistributeLearning_workflow}(b). In SFL, multiple VUs process local segments simultaneously and exchange intermediate activations with the server \cite{SFL}. Despite its parallel efficiency, SFL still suffers from substantial communication overhead due to intermediate feature transmission. Furthermore, label privacy vulnerabilities remain a critical concern depending on the configuration and loss computation placement, requiring specialized designs to protect sensitive information \cite{sl_label_attack}.
	
	\subsection{Semantic Communication: Revolutionizing the Communication Paradigm for SFL in VEI}
	\subsubsection{Deficiencies of Conventional Communication in Supporting SFL}
%	The efficacy of SFL in VEI is fundamentally constrained by the inefficiencies of conventional communication protocols when transmitting the high-dimensional intermediate features. Traditional bit-oriented communication methods exhibit several critical shortcomings. These include \textbf{excessive communication overhead} due to the indiscriminate transmission of all data bits constituting the intermediate features, irrespective of their actual relevance to the learning task, leading to significant bandwidth consumption. Furthermore, they demonstrate a \textbf{high sensitivity to network conditions}, struggling to maintain reliability and efficiency amidst the rapid channel fluctuations inherent in dynamic vehicular environments. Conventional communication has the problem to \textbf{adapt its transmission strategy to the specific requirements of the learning task} within SFL, creating a disconnect between communication processes and VEI application goals.    
	The efficacy of SFL in VEI is constrained by conventional communication protocols when transmitting high-dimensional intermediate features. Traditional bit-oriented methods exhibit critical shortcomings, including \textbf{excessive communication overhead} caused by indiscriminately transmitting all data bits regardless of their relevance to the learning task. Furthermore, they show \textbf{high sensitivity to network conditions}, struggling to maintain reliability amidst rapid channel fluctuations in vehicular environments. Finally, their inability to \textbf{adapt transmission strategies to specific learning tasks} creates a fundamental disconnect between communication processes and VEI application goals.
	
	\subsubsection{Semantic Communication for VEI}
\begin{figure*}[t]
		\centering
		\renewcommand{\figurename}{Fig.} 
		\includegraphics[width=0.85\textwidth]{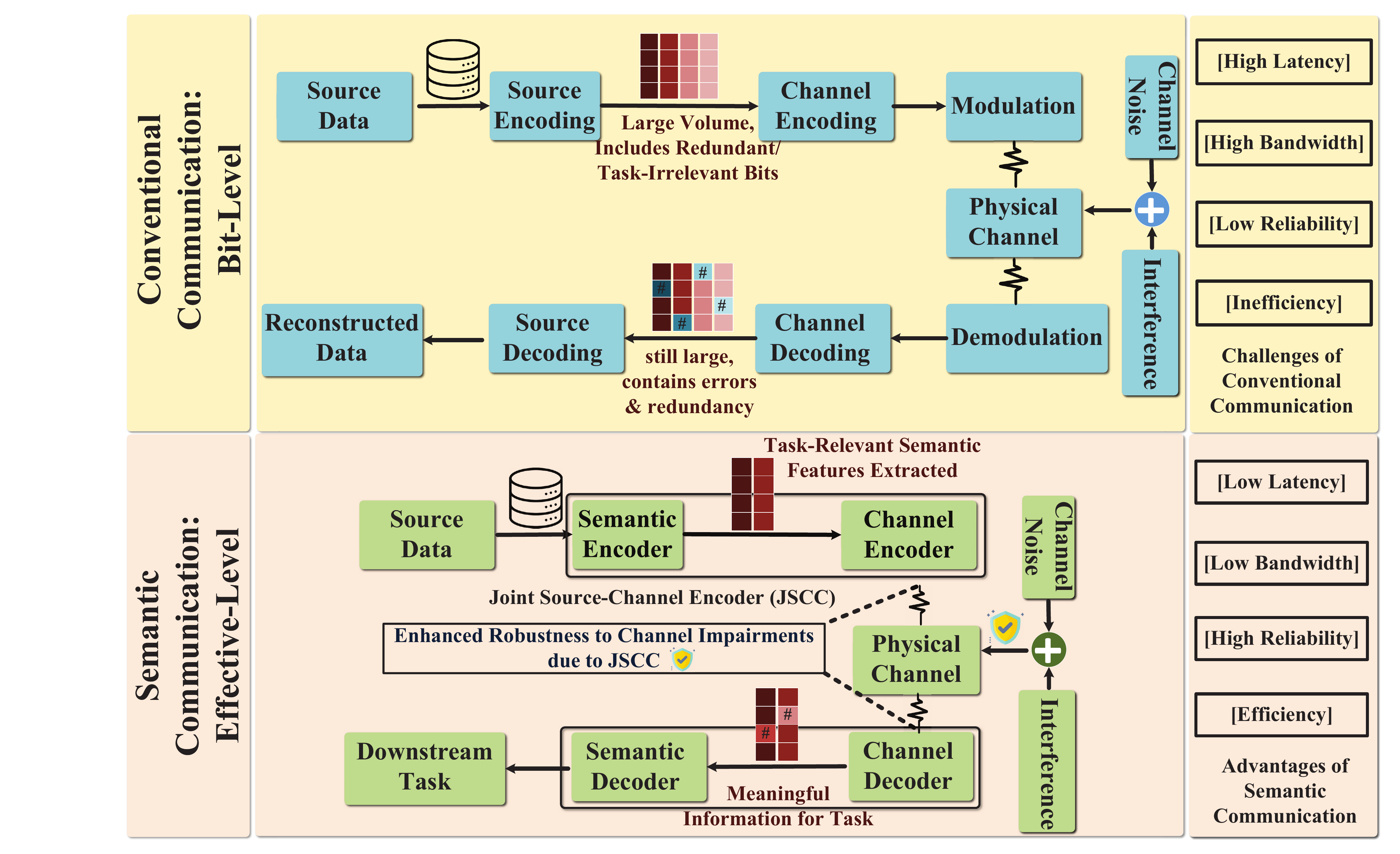}
		\caption{Comparison of Conventional Bit-Level Communication versus Semantic Effective-Level Communication pipelines.}
		\label{fig:communication_paradigm_shift}
	\end{figure*}

%	Semantic communication emerges as a transformative approach, shifting the communication paradigm from bit-perfect replication to the effective conveyance of task-relevant \textit{meaning} or \textit{semantic content}. The foundational principle is to extract and transmit only the information essential for the successful completion of an intended task at the receiver, rather than the entire raw data. This inherently allows for a substantial \textbf{reduction in communication overhead} by discarding redundant or irrelevant information. Consequently, semantic communication can significantly \textbf{enhance communication efficiency} and alleviate network congestion. Moreover, by focusing on the robust transmission of essential meaning, these systems can achieve greater \textbf{resilience to channel noise and variations}, as minor distortions at the bit level may not necessarily corrupt the conveyed semantic information. In vehicular networks, this translates to the potential for more resource-efficient collaborative perception, streamlined feature exchange for distributed model training (such as in SFL), and more reliable intelligent decision-making. The contrasting pipelines of conventional and semantic communication, highlighting these fundamental differences, are illustrated in Fig.~\ref{fig:communication_paradigm_shift}.    
	Semantic communication shifts the paradigm from bit-perfect replication to conveying task-relevant \textit{meaning} or \textit{semantic content}. By extracting and transmitting only information essential for the receiver's task, it achieves a substantial \textbf{reduction in communication overhead}. Discarding redundant data significantly \textbf{enhances communication efficiency} and alleviates network congestion. Furthermore, focusing on essential meaning provides greater \textbf{resilience to channel noise and variations}, as bit-level distortions may not corrupt the semantic content. In vehicular networks, this enables resource-efficient collaborative perception, streamlined SFL feature exchange, and reliable decision-making. Fig.~\ref{fig:communication_paradigm_shift} illustrates these contrasting pipelines.
	
	\subsubsection{Semantic Communication-empowered SFL }
	Integrating semantic communication surmounts SFL communication bottlenecks in VEI by focusing exclusively on task-relevant intermediate features. This approach mitigates excessive overhead, as semantic encoding extracts and transmits only the features vital to the learning objective, drastically reducing the data volume exchanged between clients and the server. Furthermore, employing Deep Joint Source-Channel Coding (JSCC) significantly enhances robustness against channel variations \cite{Deep-JSCC}. Through end-to-end optimization, JSCC learns compact representations that are inherently resilient to wireless noise and fading. By embedding JSCC-leveraged semantic communication into SFL, only salient, robustly encoded information is exchanged. This effectively curtails communication latency, conserves network resources, and directly aligns the transmission process with the overarching goals of VEI systems.

	\subsection{Challenges Arising from the Integration of Semantic Communication and SFL}
%	While integrating semantic communication with SFL holds considerable promise for VEI, its effective realization introduces a new set of intricate research challenges and pertinent questions that necessitate thorough investigation. Successfully addressing these challenges is paramount to unlocking the full potential of such integrated frameworks.
	While integrating semantic communication with SFL holds promise, current approaches fall short of fully addressing the unique constraints of VEI. For instance, recent semantic-enhanced frameworks \cite{FL_semantic_1}, \cite{FL_semantic_2} predominantly rely on standard FL architectures, overlooking the complexities of model splitting and label privacy. Similarly, state-of-the-art adaptive transmission schemes \cite{PADC} primarily optimize for reconstruction fidelity (e.g., PSNR), often neglecting the task-oriented objectives essential for rapid vehicular decision-making. These limitations in the existing literature give rise to a new set of intricate research challenges that necessitate thorough investigation.
	
	\subsubsection{Alignment of Semantic Extraction with Learning Tasks}
	A fundamental challenge lies in designing semantic encoder-decoder architectures that ensure the extracted semantic features maximally serve the specific learning objectives within the broader SFL framework, such as image classification or object detection performed by the global model. This necessitates robust mechanisms to effectively guide the training of semantic communication module, ensuring that the ``meaning" it preserves from the client-side features is precisely the meaning required by the server-side model segment and the ultimate artificial intelligence task. How to formulate appropriate objective functions and training strategies to achieve this task-aware semantic alignment remains a critical area of research.    
	
	\subsubsection{Quantifying and Balancing Semantic Distortion and Task Performance}
	The process of semantic compression introduces a degree of ``semantic distortion", where not all nuances of the original intermediate features might be perfectly preserved or reconstructed. It is crucial to develop effective methodologies to quantify this semantic distortion and rigorously analyze its impact on the ultimate performance of the distributed SFL model. Consequently, establishing an optimal and adaptable balance between the achieved semantic compression ratio (which inherently reflects the level of distortion) and the resultant task-specific performance (e.g., classification accuracy, reconstruction fidelity of features for the server model) is a key optimization problem that demands sophisticated solutions for SFL systems.    
	
	\subsubsection{Joint Multi-Variable Optimization in Dynamic Network Environments}
	Vehicular networks are characterized by high dynamism, including rapid channel fluctuations and varying resource availability. In this context, achieving optimal system performance for semantic-SFL requires the co-optimization of multiple interdependent variables. How to effectively and efficiently perform joint optimization of semantic communication strategies (e.g., dynamic selection of compression rates for intermediate features) and wireless communication resources (e.g., bandwidth allocation, power control for multiple SFL clients) to adapt to these real-time network variations and satisfy global system objectives—such as minimizing end-to-end latency or maximizing overall task success probability for the SFL process—remains a complex, often NP-hard, problem. This challenge calls for the development of sophisticated and computationally tractable algorithms capable of navigating this multi-dimensional optimization space.    
	
	\subsubsection{Real-time Perception and Rapid Adaptive Decision-Making}
	To enable effective dynamic optimization in semantic SFL, the integrated system must possess the capability for real-time and accurate perception of the surrounding network state for all participating VUs, encompassing parameters like channel quality, traffic load, and available computational resources. A significant research question is how to design lightweight yet effective mechanisms for network status monitoring and feedback. Furthermore, translating this perceived network intelligence into rapid and intelligent adaptive control policies for adjusting semantic compression levels for SFL feature transmission and resource allocation strategies is critical for maintaining high performance and robustness in practical, real-world deployments of semantic-enhanced SFL.    
	
	These challenges highlight the multifaceted research landscape at the intersection of semantic communication and SFL in vehicular networks, paving the way for innovative solutions such as the SC-USFL framework detailed in the subsequent chapters as a specific instantiation.   

	\section{SC-USFL: Semantic-Enhanced and Adaptive U-Shaped Split Federated Learning}
	We begin by motivating the choice of the U-SFL architecture as a specialized SFL configuration that addresses specific challenges like label privacy. Subsequently, we detail the proposed SC-USFL framework, designed as a case study to demonstrate the effective integration of semantic communication within this advanced SFL variant to tackle communication bottlenecks.
	
	\subsection{U-Shaped SFL: A Specialized SFL Architecture for Enhanced Privacy and Resource Balancing}
	\label{subsec:usfl_architecture}
	While SFL offers a flexible paradigm for distributed learning, its generic configurations can present challenges regarding the privacy of sensitive label information and optimal resource distribution. For specific scenarios like vehicular networks that handle sensitive data, more robust privacy guarantees are demanded. U-Shaped split federated learning (U-SFL) emerges as a specialized SFL architecture meticulously designed to address these nuanced requirements. The core distinction of U-SFL lies in its model partitioning strategy, which divides the deep learning model into three segments: a \textbf{head module} comprising the initial layers executed on the VU's device, a \textbf{body module} consisting of the intermediate, more computationally intensive layers offloaded to the ES, and a \textbf{tail module} that includes the final layers and is also executed on the VU's device.
	
	This U-Shaped structure is pivotal for enhancing privacy, especially for labels. By retaining the tail module on the VU's device, U-SFL ensures that sensitive label information used for final computations and loss calculation is never transmitted to the server. This creates a ``U" flow of data: from the client's head to the server's body, and then back to the client's tail for final processing. Beyond privacy, this architecture provides a balanced distribution of computational tasks, allowing VUs to handle manageable portions while offloading the heavier segment to the more powerful ES. However, despite these merits, a critical bottleneck persists in U-SFL: the communication of intermediate features. The transmission of these potentially high-dimensional features between the VU and the server over resource-constrained wireless channels imposes substantial communication overhead, underscoring the need for advanced communication techniques even within sophisticated SFL architectures.

	\subsection{The SC-USFL Framework: Architecture and Operational Workflow}
	\label{subsec:sc_usfl_framework_and_workflow}
\begin{figure}[!htbp]
		\centering
		\includegraphics[width=0.8\linewidth]{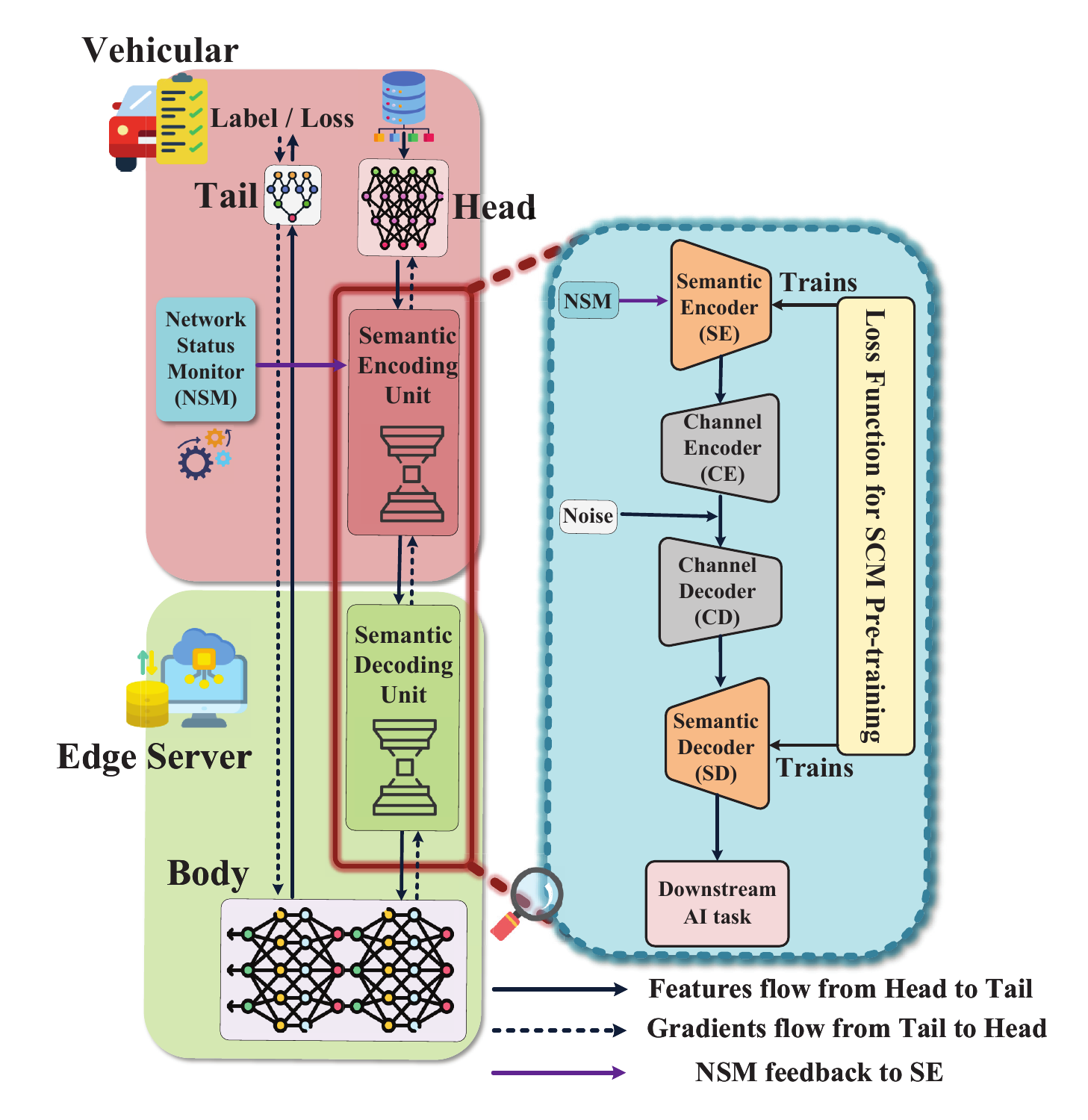}
		\caption{The SC-USFL framework: overall architecture (left) and SCM pre-training/operation details (right).}
		\label{fig:sc_usfl_architecture}
	\end{figure}

	To address the communication bottleneck in U-SFL while preserving its privacy advantages, we propose the SC-USFL framework as a specific case study. Illustrated in Fig.~\ref{fig:sc_usfl_architecture}, SC-USFL integrates task-oriented semantic communication into the U-SFL architecture.    
	
	The SC-USFL architecture retains the U-SFL's client-side head and tail modules and the server-side body module. Key additions are a client-side NSM and a SCM. The SCM consists of a semantic encoding unit (semantic encoder SE, channel encoder CE) on each VU and a semantic decoding unit (channel decoder CD, semantic decoder SD) on the ES, specifically optimizing the uplink transmission from the VU's head to the ES's body. 
	A crucial aspect is that the SCM units are pre-trained and parameter-frozen during SC-USFL training. This design choice is strategic, it eliminates the prohibitive overhead of transmitting SCM gradients over wireless uplinks, thereby preserving communication efficiency, and stabilizes the distributed training process against channel-induced gradient variance.
%	A crucial aspect is that the SCM units are pre-trained end-to-end for a relevant downstream AI task and their parameters are frozen during SC-USFL training, ensuring efficient semantic compression without ongoing SCM update overhead.    
	
	The operational workflow per communication round (Fig.~\ref{fig:sc_usfl_architecture}) is as follows: 
%	\begin{enumerate}
%		\item \textbf{VU-Side Head Processing:} The VU's Head module processes local raw data into intermediate features. 
%		\item \textbf{Uplink Semantic Transmission:} These features are compressed by the VU's pre-trained semantic encoding unit (informed by the NSM for adaptivity) and transmitted to the ES. 
%		\item \textbf{ES-Side Processing:} The ES's pre-trained semantic decoding unit reconstructs the features, which are then processed by the Body module. 
%		\item \textbf{Downlink Feature Transmission:} Resultant features from the ES's Body are sent back to the VU's Tail module directly. 
%		\item \textbf{VU-Side Tail Processing and Local Update:} The VU's Tail module performs final computations, calculates loss using local private labels (preserving label privacy), computes gradients, and updates its Head and Tail modules. 
%	\end{enumerate}
	\begin{enumerate}
		\item \textbf{VU-Side Extraction:} The VU's head module extracts intermediate feature maps from local raw data.
		\item \textbf{Uplink Semantic Transmission:} Guided by the NSM, the semantic encoding unit compresses features into lower-dimensional semantic vectors and maps them to analog symbols for robust wireless transmission.
		\item \textbf{ES-Side Reconstruction \& Processing:} The ES's semantic decoding unit receives the noisy analog signals and decodes them to reconstruct the semantic feature maps. These reconstructed features are then fed into the server-side body module for forward propagation.
		\item \textbf{Downlink Transmission:} Resultant features from the body module are transmitted back to the VU.
		\item \textbf{Local Update:} The VU's tail module computes loss using private labels to update local models, ensuring label privacy.
	\end{enumerate}
	This process occurs in parallel for all VUs, with the ES aggregating body module updates. The SCM's primary role is to enhance uplink efficiency.    
	
	\subsection{Adaptive Semantic Compression for Dynamic Networks}
	\label{subsec:adaptive_semantic_compression}
	Furthermore, the SC-USFL framework incorporates an adaptive mechanism through the NSM module. The NSM module plays a critical role by continuously monitoring real-time wireless channel conditions (e.g., Signal-to-Noise Ratio (SNR), channel fading characteristics). This crucial channel state information (CSI) is then fed back to the VU's semantic encoding unit, specifically influencing its SE component. Armed with this real-time CSI, the semantic encoding unit can intelligently adjust its semantic compression strategy. For instance, it can dynamically select an appropriate compression ratio (CR) from a predefined set to best suit the current network conditions. This adaptive capability allows SC-USFL to navigate the fundamental trade-off between communication efficiency and task performance \cite{PADC}. 
	Notably, utilizing passive monitoring and constant-time complexity logic, the NSM incurs negligible overhead, ensuring practical real-time viability.
	The ability to dynamically adjust the semantic compression rate based on real-time network perception is a key design principle of SC-USFL, enabling robust and efficient operation in the face of unpredictable vehicular network dynamics. 
	
	The U-Shaped topology strategically optimizes the trade-off among latency, energy, accuracy, and privacy. We employ a shallow split to minimize on-device energy, utilizing the integrated SCM to compress intermediate features and mitigate the communication latency typically associated with shallow offloading. Furthermore, offloading the computation-heavy backbone maximizes accuracy, while the fixed local tail structurally guarantees label privacy by isolating ground truths on the VU.

	\section{Performance Evaluation}
	\subsubsection{Experiment Setting}
	To validate the efficacy of our proposed SC-USFL framework, we conduct a comprehensive case study simulating a vehicular edge intelligence scenario where multiple VUs collaboratively train an image classification model with an edge server (ES). 
	The experiments are performed on the widely used CIFAR10 dataset. 
	Our SC-USFL framework employs a ResNet-50 based head module and a classification tail on each VU, while the ES utilizes a ViT-B/16 for body module computations. 
%	The core Semantic Communication Module (SCM), based on a DeepJSCC architecture, is pretrained on CIFAR-10 and its parameters are frozen during SC-USFL training.
	The core SCM, based on a Deep JSCC architecture, is pre-trained end-to-end as an autoencoder under simulated AWGN channel. Its parameters are subsequently frozen during the distributed SC-USFL training phase to decouple feature extraction from channel robustness. We compare SC-USFL against several baselines: traditional FL, USFL without semantic compression, centralized training, and local training. 
	Key training parameters include 200 communication rounds, 3 local training epochs before aggregation, a batch size of 64, and the Adam optimizer with a learning rate of 0.0001. 
	CRs are selected from the set comprising 1/3, 1/6, 1/8, and 1/12. This specific selection aligns with foundational benchmarks \cite{Deep-JSCC} to ensure fair comparison and guarantees engineering feasibility, as these fractions map accurately to integer channel dimensions in the neural network. Wireless channels are modeled as both additive white Gaussian noise (AWGN) and Rayleigh fading channels to assess robustness. 
	Performance is evaluated based on test accuracy, training loss, per-round latency, and task success probability.

	\subsubsection{Performance Analysis}
\begin{figure}[htbp!] 
		\centering % 整体居中
		\subfloat[ Test Accuracy under AWGN ]{% 子图(a)的标题
			\includegraphics[width=0.8\columnwidth]{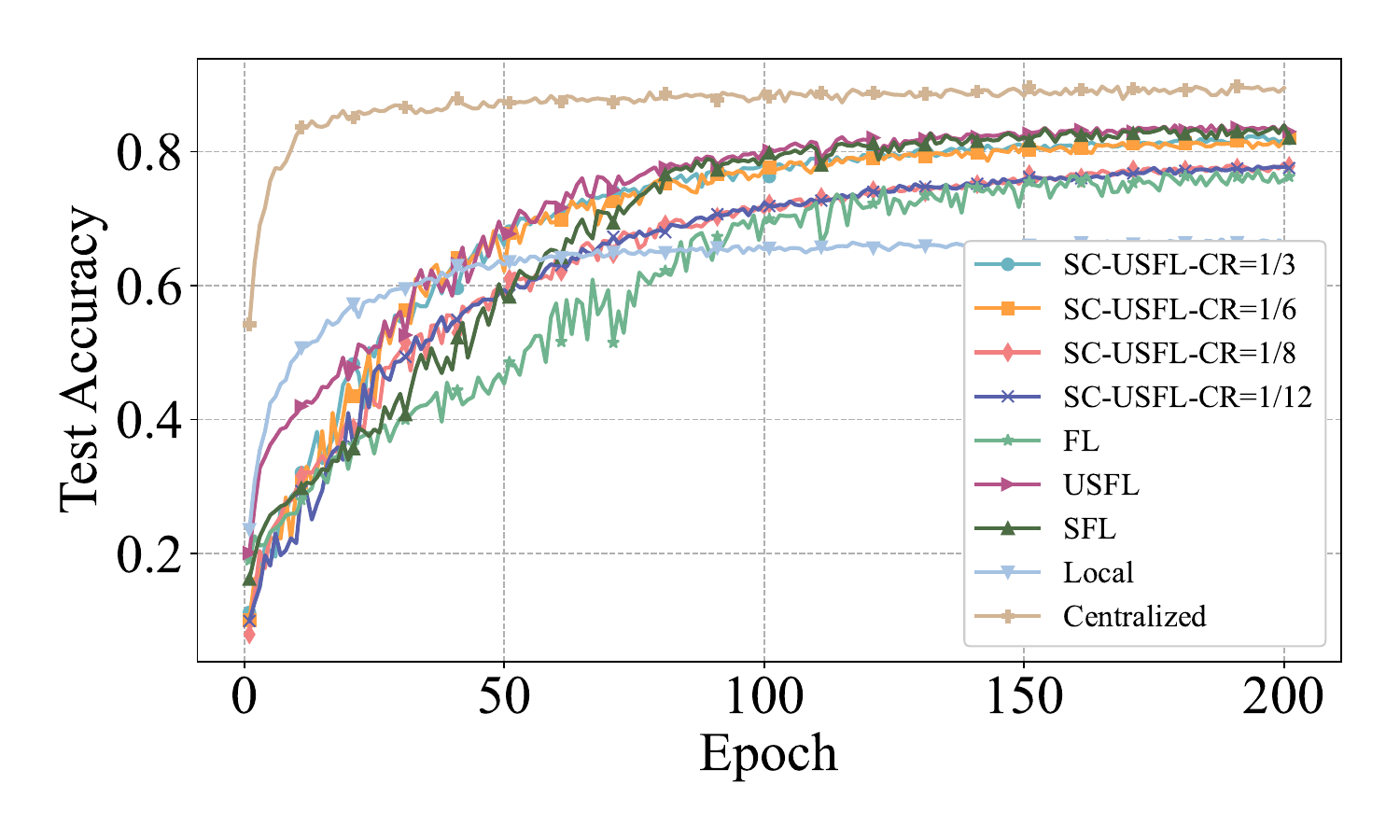}
			\label{fig:test_accuracy_awgn} % 子图(a)的新标签
		}
		% 在两个子图之间添加一些垂直间距，可以根据需要调整
		\vspace{1ex} % 可选的垂直间距，1ex 大约是一个字母'x'的高度
		\subfloat[ Total Latency Comparison ]{% 子图(b)的标题
			\includegraphics[width=0.8\columnwidth]{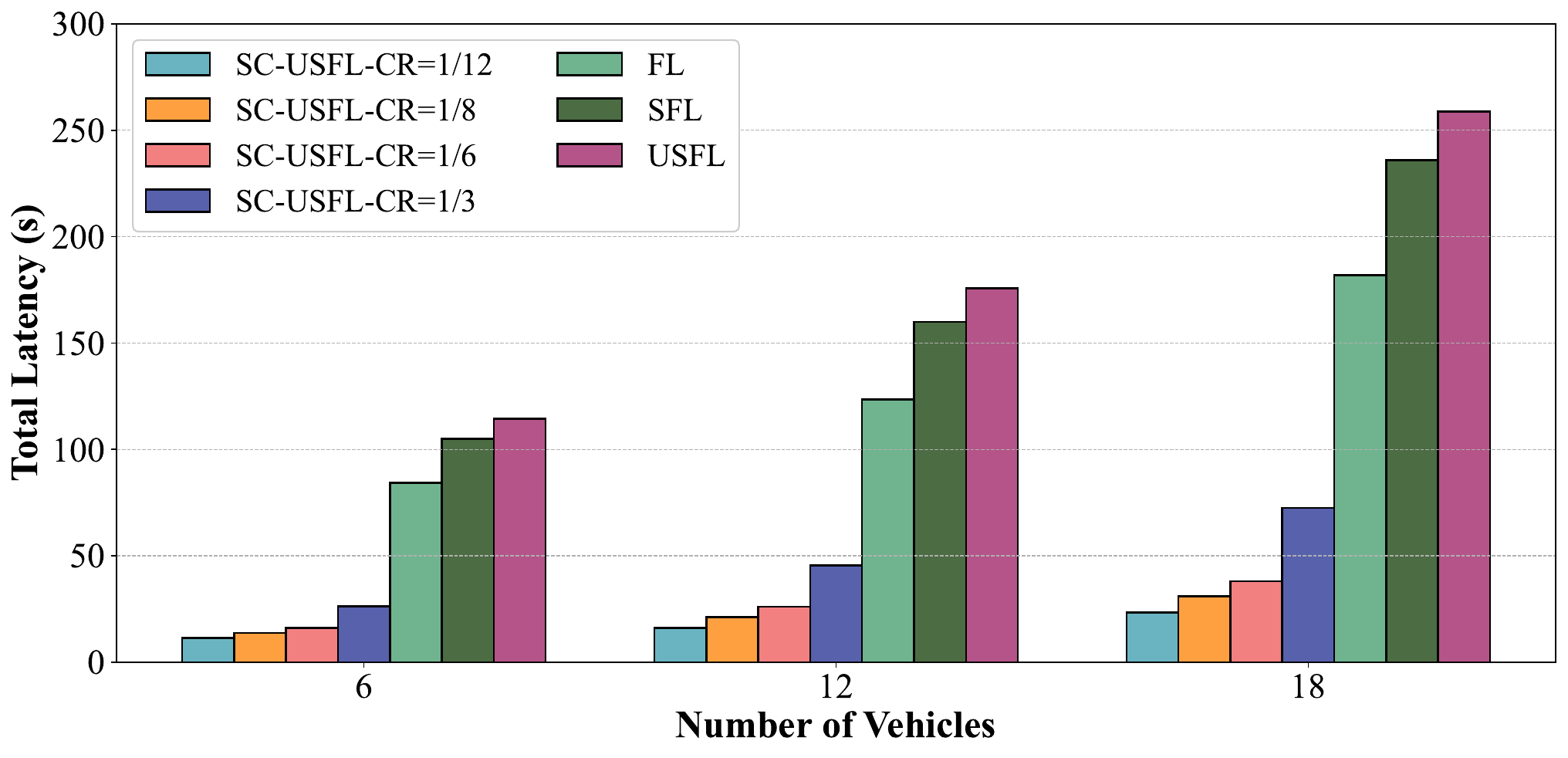}
			\label{fig:total_latency_comparison} % 子图(b)的新标签
		}
		\caption{ Performance comparison of SC-USFL with baseline architectures. (a) Test accuracy under AWGN channel. (b) Per-round total latency. }
		\label{fig:performance_comparison_stacked_main} % 整个大图的新标签
	\end{figure}

%	Comprehensive analysis considering both task accuracy as shown in Fig.~\ref{fig:test_accuracy_awgn} and communication latency as depicted in Fig.~\ref{fig:total_latency_comparison} reveals the significant advantages of the SC-USFL framework in vehicular edge intelligence. 
%	By introducing task-oriented semantic compression, SC-USFL achieves a substantial reduction in communication latency and effectively addresses the communication pressure arising from an increasing number of vehicles, all while maintaining competitive learning task performance. 
%	Compared to traditional Federated Learning (FL), FL experiences sharply escalating communication latency as the number of vehicles increases and typically imposes higher computational demands on vehicular units. 
%	Standard U-Shaped Split Federated Learning (USFL), while alleviating computational burdens, still suffers from significant communication latency growth with more vehicles due to the transmission of uncompressed intermediate features, which becomes a performance bottleneck. 
%	SC-USFL, through its semantic compression mechanism, trades a minor task performance variation for considerable gains in communication efficiency. 
%	This inherent flexibility in balancing communication efficiency and task accuracy positions SC-USFL as a more promising and practical distributed learning solution for resource-constrained and dynamic vehicular network environments compared to FL and USFL, especially for applications requiring a delicate balance between performance and real-time responsiveness.
	Comprehensive analysis of task accuracy (Fig.~\ref{fig:test_accuracy_awgn}) and communication latency (Fig.~\ref{fig:total_latency_comparison}) reveals the significant advantages of SC-USFL. 
	Compared to traditional FL, which imposes high computational and communication burdens, SL approaches offer relief. 
	However, both the standard SFL \cite{SFL} and USFL baselines, despite achieving high accuracy, suffer from prohibitive communication latency as the vehicle count increases. This is due to the transmission of uncompressed high-dimensional intermediate features, which becomes a severe bottleneck. 
	In contrast, SC-USFL leverages task-oriented semantic compression to trade a minor accuracy variation for substantial gains in communication efficiency. 
	This flexibility positions SC-USFL as a more practical solution for resource-constrained vehicular networks compared to FL, SFL, and USFL.
	
\begin{figure}[htbp!] 
		\centering % 整体居中
		\subfloat[AWGN Channel.]{% 子图(a)的标题
			\includegraphics[width=0.7\columnwidth]{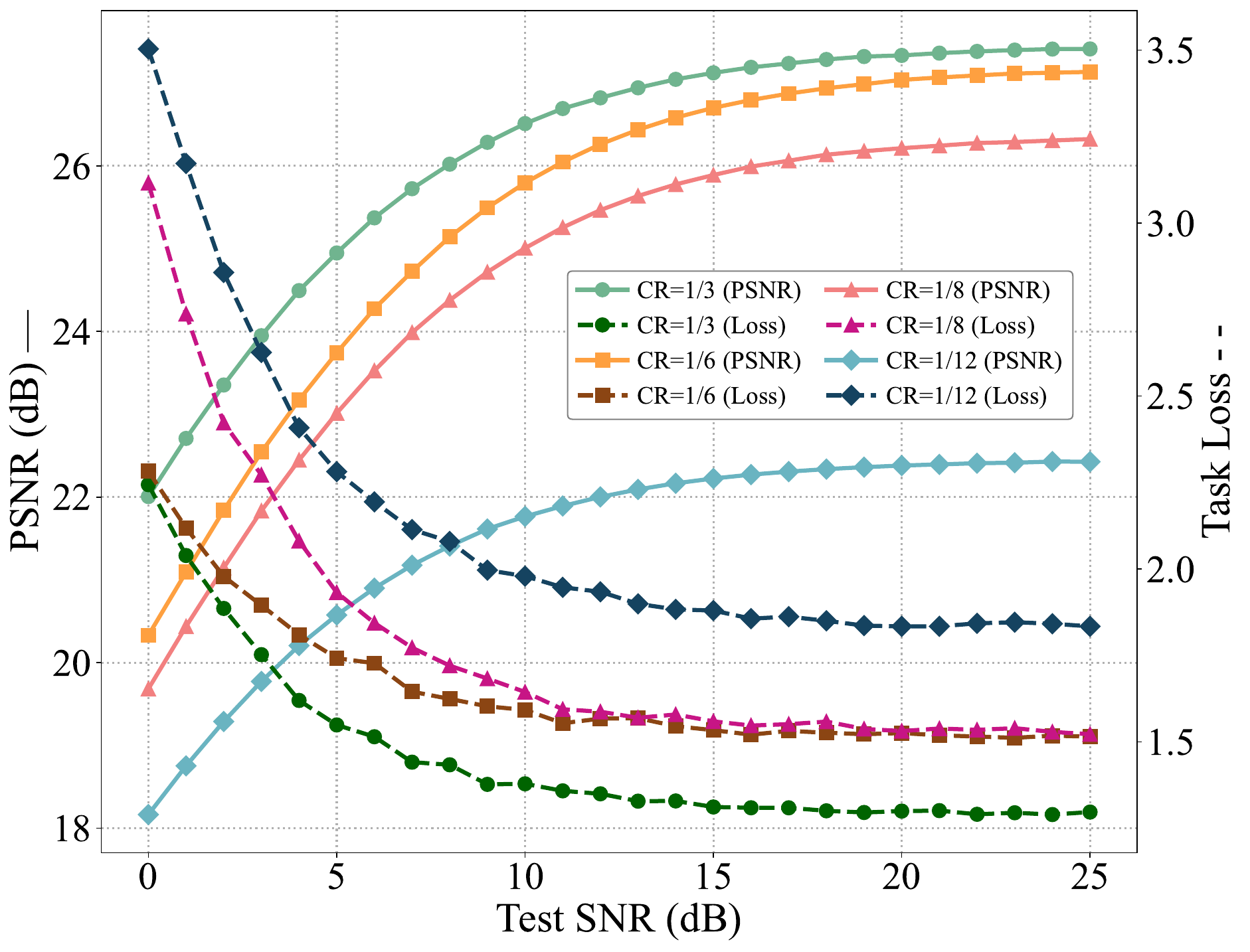} % 稍微调整宽度以便标题显示
			\label{fig:psnr_taskloss_awgn} % 子图(a)的标签
		}
		% 在两个subfloat之间加一个空行，或者使用 \\ 来强制换行
		\vspace{1ex} % 可选：在子图间增加一些垂直间距
		\subfloat[Rayleigh Channel.]{% 子图(b)的标题
			\includegraphics[width=0.7\columnwidth]{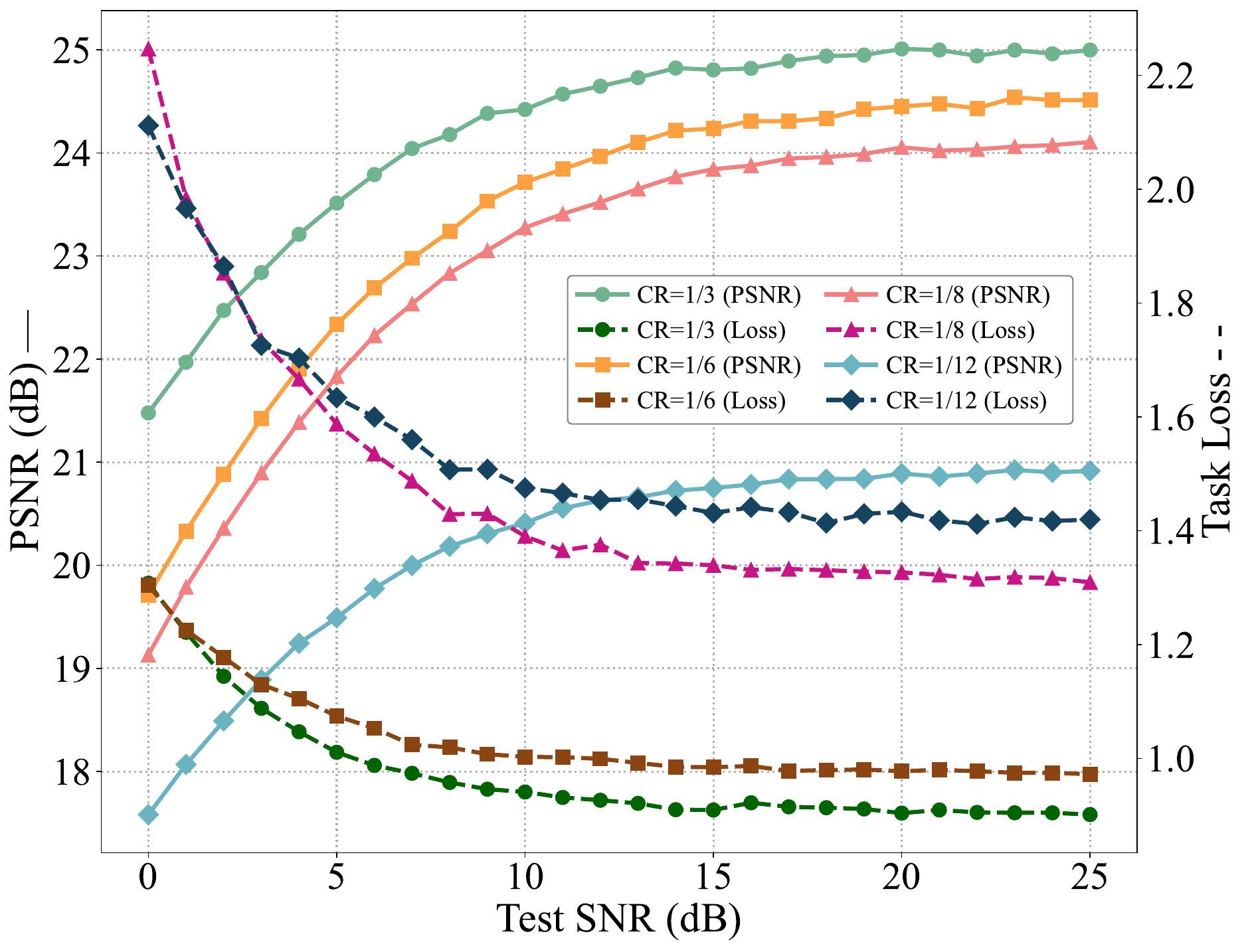} % 稍微调整宽度以便标题显示
			\label{fig:psnr_taskloss_rayleigh} % 子图(b)的标签
		}
		\caption{ Robustness and performance trade-offs of SC-USFL under varying test SNRs and compression ratios for AWGN and Rayleigh channels. Higher CR values (e.g., 1/3) represent milder compression.  }
		\label{fig:robustness_tradeoff}
	\end{figure}

	The results from Fig.~\ref{fig:psnr_taskloss_awgn} and Fig.~\ref{fig:psnr_taskloss_rayleigh} clearly demonstrate a core trade-off in semantic communication within SC-USFL: stronger compression (lower CR), while beneficial for reducing data transmission volume, can lead to higher semantic distortion, thereby affecting reconstruction quality (lower PSNR) and ultimate task performance (higher task loss). The consistent behavior observed across different CRs and SNRs, even under the challenging Rayleigh fading channel, further validates the robustness and effectiveness of the SC-USFL framework's design. This also underscores the importance of dynamically adjusting the compression rate based on real-time channel conditions and task requirements (i.e., adaptive compression).
	
	While the simulation results validate the effectiveness of SC-USFL, several limitations warrant note. 
	Regarding adaptivity, the NSM currently operates on a discrete action space (i.e., fixed compression ratios). This coarse-grained adaptation may induce quantization inefficiencies compared to continuous rate control.
	Regarding robustness, our evaluation assumes perfect CSI. In practice, high-mobility-induced CSI aging and estimation errors may degrade the precision of the NSM's decisions.
	Finally, regarding scope, this work primarily addresses the transmission bottleneck of visual data. The integration of heterogeneous modalities  requires designing synchronized multi-modal semantic encoders, which remains a critical direction for future extension.
	
	\section{Open Research Directions}
	\label{sec:open_research_directions}
	
	The integration of semantic communication with SFL opens several avenues for future research. Addressing these directions can further unlock the potential of intelligent vehicular systems.
	
	\subsection{Generalizability Across Tasks and Modalities}
	\label{subsec:generalizability}
	Cross-Task Semantic Representation: Designing SCMs adaptable to diverse tasks remains a challenge. Future studies should investigate foundation models (e.g., large-scale pre-trained transformers) to extract universal semantic features that can be fine-tuned for object detection or trajectory prediction with minimal overhead, replacing task-specific encoders. 
	
	Multi-Modal Semantic Communication: Vehicular intelligence relies on fusing heterogeneous data. A vital direction is developing cross-modal attention mechanisms to efficiently align and compress unstructured LiDAR point clouds with structured RGB images, capturing a holistic environmental understanding.
	
	\subsection{Security and Privacy in Semantic Communication}
	\label{subsec:security_privacy_semantic}
	Defenses against Model Inversion: While semantic compression offers obfuscation, model inversion remains a risk. Future research should rigorously quantify privacy bounds using mutual information estimation and develop semantic differential privacy mechanisms to provide formal guarantees against feature reconstruction attacks. 
	
	Robustness against Adversarial Perturbations: Semantic encoders introduce new attack surfaces. Specific studies are needed to design certified robustness verification methods and active adversarial training strategies that immunize semantic encoders against targeted gradient-based perturbations without compromising transmission efficiency.
	
	\subsection{Semantic Knowledge Management and Utilization}
	\label{subsec:semantic_knowledge_management}
	Distributed Knowledge Bases: Efficiently managing knowledge is key. Potential studies could utilize graph neural networks (GNNs) to maintain and align distributed semantic knowledge graphs among vehicles, facilitating efficient semantic reasoning and discovery protocols. 
	
	Context-Aware Communication: Leveraging shared context can reduce redundancy. Future work should explore conditional entropy coding, where the client transmits only the ``semantic residuals'' relative to the server's existing knowledge base, drastically reducing overhead.
	
	\subsection{Information Freshness and Value in Semantic Transmissions}
	\label{subsec:information_freshness_value}
	Semantic Age of Information (SAoI): Beyond packet timestamps, research should formulate SAoI metrics that quantify the freshness of the ``meaning'' itself, prioritizing features that reduce model uncertainty \cite{SAoI_1}. 
	
	Value of Semantic Information (VoSI): It is critical to quantify the intrinsic value of features. We suggest employing shapley value analysis to measure the contribution of specific semantic packets to the global model's accuracy, enabling importance-aware resource scheduling \cite{SAoI_2}.

	\section{Conclusion}
	\label{sec:conclusion}
	This paper tackled communication inefficiency and privacy challenges in distributed learning for VEI.  While SFL is a promising distributed approach, it often faces high communication overhead and potential label privacy issues. We posited semantic communication as a key paradigm to address SFL's communication burdens by transmitting only task-relevant meaning.
	Our novel contribution, the SC-USFL framework, serves as a compelling case study. It integrates semantic communication with U-SFL, an SFL architecture inherently strong on label privacy. SC-USFL features a SCM with pre-trained, parameter-frozen units for efficient uplink semantic compression, and a NSM for real-time adaptive compression. Evaluations confirmed SC-USFL effectively balances communication load and task performance while preserving U-SFL's privacy benefits.
	This research offers SC-USFL as a practical VEI solution and, more broadly, highlights the significant potential of integrating semantic communication with SFL, openning up several promising avenues for future exploration.
	
	\section*{Acknowledgment}
	This work is partly supported by the Fundamental and Interdisciplinary Disciplines Breakthrough Plan of the Ministry of Education of China under Grant No. JYB2025XDXM116, NSF of Sichuan under Grant No. 2025YFHZ0093 and by Horizon European Union Grant No. 101086159 and No. 101131117.

\end{document}